\documentclass[10pt,twocolumn,letterpaper]{article}
\usepackage{cvpr}
\usepackage{times}
\usepackage{float}
\usepackage[utf8]{inputenc}
\usepackage[T1]{fontenc}

\usepackage{url}
\usepackage{booktabs}
\usepackage{amsfonts}
\usepackage{nicefrac}
\usepackage{microtype}
\usepackage{epsfig}
\usepackage{graphicx}
\usepackage{amsmath}
\usepackage{amssymb}
\usepackage{multirow}
\usepackage{multicol}
\usepackage{array}
\usepackage{algorithm,algorithmicx,algpseudocode}

\newcommand\ignore[1]{}
\newcommand\mydots{\hbox to 0.7em{.\hss.\hss.}}
\renewcommand\lll{l_1,\mydots,\thinspace l_d}
\newcommand\plll{p(\lll)}
\newcommand\pli{p^c(l_i|I)}
\newcommand\cwdh{$cw\Delta H$}
\newcommand\cwmidh{$cw-Image\Delta H$}
\newcommand\cwdhmath{cw\Delta H}
\newcommand\tabbf[1]{\textbf{0.#1}}
\renewcommand\eqref[1]{Eq.~(\ref{#1})}

\usepackage[pagebackref=true,breaklinks=true,letterpaper=true,colorlinks,bookmarks=false]{hyperref}
\cvprfinalcopy

\def\cvprPaperID{****} 

\ifcvprfinal\fi

\title{Informative Object Annotations \\ \emph{Tell Me Something I Don't Know}}
\author{
\textbf{Lior Bracha}\\
Bar-Ilan University\\
{\tt\small lior.bracha@live.biu.ac.il}
\and
\textbf{Gal Chechik}\\
Bar-Ilan University, NVIDIA Research\\
{\tt\small gal.chechik@biu.ac.il}
}

\begin{document}
\maketitle
\begin{abstract}
    Capturing the interesting components of an image is a key aspect of image understanding. When a speaker annotates an image, selecting labels that are informative  greatly depends on the prior knowledge of a prospective listener. 
    Motivated by cognitive theories of categorization and communication, we present a new unsupervised approach to model this prior knowledge and quantify the informativeness of a description. Specifically, we compute how knowledge of a label reduces uncertainty over the space of labels and utilize this to rank candidate labels for describing an image. While the full estimation problem is intractable, we describe an efficient algorithm to approximate entropy reduction using a tree-structured graphical model. We evaluate our approach on the open-images dataset using a new evaluation set of 10K ground-truth ratings and find that it achieves $\sim65$\% agreement with human raters, largely outperforming other unsupervised baseline approaches.
\end{abstract}

\begin{figure}[t]
\begin{center}
    \includegraphics[width=1\linewidth]{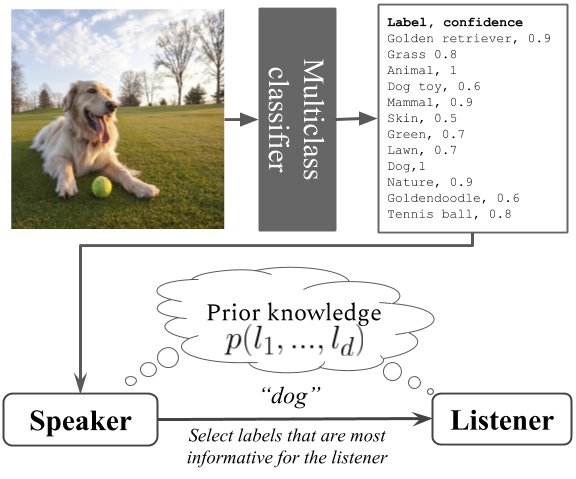}
\end{center}
  \caption{ \textbf{The problem of informative labeling}. An image is automatically annotated with multiple labels. A "speaker" is then given these labels and their confidence scores and has to select $k$ labels to transmit to a listener, such that the listener finds them informative given her prior knowledge. The prior knowledge is assumed to be common to both the speaker and the listener.}
\label{fig:key}
\end{figure}

\section{Introduction}
How would you label the photo in Figure 1? If you answered "a dog", your response agrees with what most people would answer. Indeed, people are surprisingly consistent when asked to describe what an image is "about" \cite{rosch1976basic}. They intuitively manage to focus on what is ``informative" or "relevant" and select terms that reflect this information. In contrast, automated classifiers can produce a large number of labels that might be technically correct, but are often non-interesting.

\begin{figure*}[ht]
    \begin{center}
        \includegraphics[width=1\linewidth]{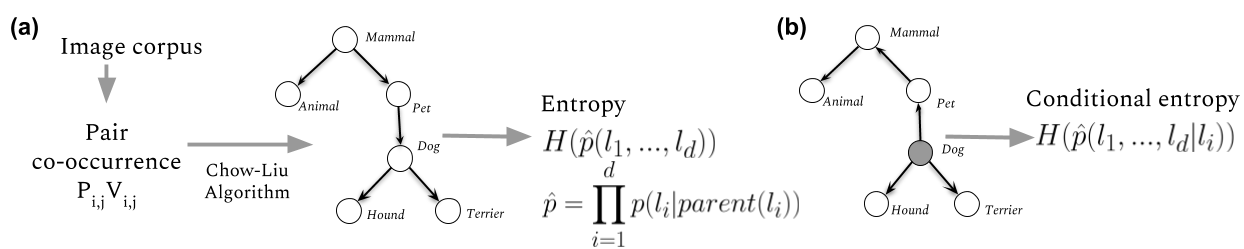}
    \end{center}
    \caption{\textbf{Uncertainty over labels can be estimated through measuring the entropy of its joint distribution, and computed efficiently using a tree-structured probabilistic graphical model (PGM)}. \textbf{(a)} An image corpus is used for collecting pairwise label co-occurrence. Then, a tree-structured graphical model is learned using the Chow-Liu algorithm. Computing the entropy of the approximated distribution $\hat{p}$ has a run-time that is linear in the number of labels. \textbf{(b)} To compute the entropy conditioned on a label $l_{dog} = true$, the marginal of that node is set to [0,1]. Then, the graph edges are redirected and rest of the distribution is updated using the conditional probability tables represented on the edges. Finally, we compute the entropy of the resulting distribution. }
    \label{fig:method}
\end{figure*}

A natural approach to ascertain importance lies in the context of the specific task. For instance, classifiers can be efficiently trained to identify dog breeds or animal species. More generally, each task defines importance through a supervision signal provided to the classifier \cite{andreas2016reasoning,vedantam2017context,luo2018discriminability}. Here we are interested in a more generic setup, where no down-stream task dictates the scene interpretation. This represents the challenge that people face when describing a scene to another person, without any specific task at hand.

The principles that govern informative communication have long been a subject of research in various fields from philosophy of language and linguistics to computer science. In the discipline of pragmatics, Grice's maxims state that ``one tries to be as informative as one possibly can." \cite{grice1975logic}. But the question remains, ``Informative about what?" How can we build a practical theory of informative communication that can be applied to concrete problems with real-world data?

In this paper, we address the following concrete learning setup (Figure \ref{fig:key}). A speaker receives a set of labels predicted automatically from an image by a multiclass classifier. It also receives the confidence that the classifier assigns to each prediction. Then, it aims to select a few labels to be transmitted to a listener, such that the listener will find those labels informative. The speaker and listener also share the same prior knowledge in the form of the distribution of labels in the image dataset.

We put forward a quantitative theory of how speakers select terms to describe an image. The key idea is that communicated terms are aimed to reduce the uncertainty that a listener has about the semantic space. We show how this "theory-of-mind" can be quantitatively computed using information-theoretic measures. In contrast with previous approaches that focused on visual aspects and their importance \cite{elazary2008interesting,berg2012understanding,spain2011measuring,liu2011learning,borji2015salient}, our measures focus on information about the semantics of labels. 

To compute information content of a label, we build a probabilistic model of the full label space and use it to quantify how transmitting a label reduces uncertainty. Specifically, we compute the entropy of the label distribution as a measure of uncertainty, and also quantify how much this entropy is reduced when a label is set to be true. 

Importantly, computing these measures over the full distribution of labels is not feasible because it requires to aggregate an exponentially-large set of label combinations. We show how the entropy and other information theoretic measures can be computed efficiently by approximating the full joint distribution with a tree-structured graphical model (a Chow-Liu tree). We then treat entropy-reduction as a scoring function that allows us to rank all labels of an image, and select those that reduce the entropy most. We name this approach IOTA, for \textit{Informative Object Annotations}. \vspace{5pt}

We test this approach on a new evaluation dataset: 10K images from the open-images dataset \cite{OpenImages} were annotated with informative labels by three raters each. We find that human annotations are in strong agreement ($\sim\!70\%$) with the uncertainty-reduction measures, just shy of inter-rater agreement and superior to 4 other unsupervised baselines. \vspace{5pt}

Our main contributions are as follows: (1) We describe a novel learning setup of selecting important labels without direct supervision about importance. (2) We develop an information-theoretic framework to address this task, and propose scoring functions that can be used to solve it. (3) We further describe an efficient algorithm for computing these scoring functions, by approximating the label distribution using a tree-structured graphical model. (4) We provide a new evaluation set of ground-truth importance ratings based on 10K images from the open-images dataset. (5) We show that IOTA achieves high agreement with human judgment on this dataset.

\section{Related work}
\paragraph*{Image importance and object saliency.}
The problem of deciding which components in an image are important has been studied intensively. The main approaches involved identifying characteristics of objects and images that could contribute to importance, and use labeled data for predicting object importance. Elazary and Itti \cite{elazary2008interesting} considered the order of object naming in the LabelMe dataset \cite{russell2008labelme} as a measure of the interest of an object and compare that to salient locations predicted by computational models of bottom-up attention. The elegant work of Spain and Perona \cite{spain2011measuring} examined which factors can predict the order in which objects will be mentioned given an image. Berg et al. \cite{berg2012understanding} characterized factors related to semantics, to composition and to the likelihood of attribute-object, and investigated how these affected the measures of importance. \cite{Ordonez_2013_ICCV} focused on predicting entry-level classes using a supervised approach. These studies also make it clear that the object saliency is strongly correlated with its perceived importance \cite{liu2011learning,borji2015salient}. 

These studies differ from the current work in two significant ways. First, they largely focus on visual properties of objects in images, while our current approach focuses on modeling the labels structure, and only uses image-based information in the form of label confidence as predicted by a classifier. Second, these work largely take a supervised approach using measures of importance in a training set to build predictive models of label importance. In contrast, IOTA, our approach is unsupervised, in the sense that our model is not directly exposed to labeled information about object importance.

\noindent
\textbf{Information theory and measures of relevance.}
The problem of extracting informative components from a complex signal was studied from an information-theoretic perspective through the information bottleneck principle (IB) \cite{tishby2000information, chechik2005information,zaslavsky2018efficient}. In contrast to the current work, in IB, a signal, $X$, is compressed into $T$ such that it maximizes information about another variable $Y$, that can be viewed as a supervision variable.

\noindent
\textbf{Pragmatics, Relevance theory}
In pragmatics, effective communication has been characterized by the \textit{cooperative principle} \cite{grice1975logic}, which views communication as a cooperative interaction between a speaker and a listener. These principles were phrased in Grice's maxims, stating that ``one tries to be as informative as one possibly can" and ``and does not give information that is false or that is not supported by evidence". Our approach provides a concrete quantitative  realization to these principles. Inspired by Grice's work, Sperber and Wilson proposed a framework called relevance theory \cite{wilson2002relevance,wilson2012meaning}. This theory highlighted that a speaker provides cues to a listener, who then interprets them in the context of what she already knows and what the speaker may intended to transmit. 

\section{Our approach}
The key idea of our approach is to quantify the relevant-information content of a message, by modelling what the listener does not know, and find labels that reduce this uncertainty. To illustrate the idea, consider a label that appears in most of the images in a dataset (e.g., \textit{nature}). If the speaker selects to transmit that label, it provides very little information to the listener, because they can already assume that a given image is annotated with that label. In contrast, if the speaker transmits a label that is less common, appearing in only half of the images, more of the listener's uncertainty would be removed.

A more important property of multi-label uncertainty is that labels are \textit{interdependent}:  transmitting one label can reduce the uncertainty of others. This property is evident when considering label hierarchy, for example, \textit{golden-retriever} = true implies that \textit{dog} = true. As a result, transmitting a fine-grained label removes more entropy than a more general label. This effect however, is not limited to hierarchical relations. For instance, because the label \textit{street} tends to co-occur with \textit{car} and other vehicles, transmitting \textit{street} would reduce the overall uncertainty by reducing uncertainty in correlated co-occurring terms. 

Going beyond these examples, we aim to calculate how a revealed label affects the listener's uncertainty. 
For this purpose, the Shannon entropy is a natural choice to quantify uncertainty, pending that we can estimate the prior joint distribution of labels. 
Clearly, modelling the entire prior knowledge about the visual world of a listener is beyond our current reach. Instead, we show how we can approximate the entire joint distribution by building a compact graphical model with a tree structure. This allows us to efficiently compute properties of the joint distribution over labels and more specifically, estimate listener uncertainty and label-conditioned uncertainty. 

We start by describing an information-theoretic approach for selecting informative labels by estimating uncertainty and label-conditioned uncertainty. We then describe an algorithm to effectively compute these quantities in practice.

\subsection{The problem setup}
Assume that we are given a corpus of images, each annotated with multiple labels from a vocabulary of $d$ terms ${\cal{L}}=(l_1, \ldots, l_d)$. Since we operate in a noisy labeling setup, we treat the labels as binary random variables $l_i \in \{true, false\}$. We also assume that for each image $I$, labels are accompanied with a score reflecting the classifier's confidence in that label $\pli$.\footnote{Such confidence scores can be obtained from classifier predictions, assuming that these confidence scores are calibrated, namely, reflect the true fraction of correct labels. In practice, many large-scale models indeed calibrate their scores, as we discuss in the experimental section.} The goal of the speaker is to select $k$ labels to be transmitted to the listener, such that they are most "useful" or informative. 

\subsection{Information-theoretic measure of importance} \label{sec:scoring_functions}
Let us first assume that we can estimate the distribution over labels that a listener has in mind. Clearly, this is a major assumption, and we discuss below how we relax this assumption and approximate this distribution.

Given this distribution, we wish to measure the uncertainty it reflects, as well as how much this uncertainty is reduced when the speaker reveals a specific label. A principled measure of the uncertainty about random variables is the Shannon entropy of their joint distribution $H(L_1,\ldots,L_d)$ \cite{cover2012elements}. We use a notation that makes it explicit that the entropy depends on the distribution, where the entropy is defined as 
\begin{equation}
\label{eq:H-naive}
    H[\plll] \!=\! -\!\!\!\sum_{\lll} \plll \log \plll.
\end{equation}
Here, summation is over all possible assignments of the $d$ labels, an exponential number of terms that cannot be computed in practice. We show below how to approximate it. 

The amount of entropy that is reduced when the speaker transmits a subset of the labels ${\cal{L}'} = \{l_i, l_j, l_k, \ldots\}$, is

\begin{equation*}
    \label{eq:delta-H-subset}
    \Delta H({\cal L}') = H[\plll] - H[p(\lll|{\cal L}'=true)]\quad,
\end{equation*}

where ${\cal L'}=true$ means that all labels in ${\cal L'}$ are assigned a true value. For simplicity, we focus here on the case of transmitting a single label $l_i$ (see also \cite{deweese1999measure}), and define the \textbf{\textit{per-label entropy-reduction}}
\begin{equation}
    \label{eq:delta-H}
    \Delta H(i) = H[\plll] - H[p(\lll|l_i\!\!=\!\!true)].
\end{equation}

This measure has several interesting properties. It has a similar form to the Shannon mutual information, $MI(X;Y)=H(X)-H(X|Y)$, which is always positive. However, the condition on the second term is only over a single value of the label ($l_i=true$). As a result, \eqref{eq:delta-H} can obtain both negative and positive values. When the random variables are independent, $\Delta H(i)$ is always positive, because the entropy can be factored using the chain rule, and obeys $H(L_1,\ldots,L_d) - H(L_1,\ldots,L_d|L_i) = \sum_{j\ne i} H(L_j)>0$ (Sec 2.5 \cite{cover2012elements}). However, when the variables are not independent, collapsing one variable to a True value can actually increase the entropy of other co-dependant variables. As an intuitive example, the base probability of observing a lion in a city is very low, and has low entropy. However, once you see a sign ``zoo", the entropy of facing a lion rises. 

The second important property of $\Delta H(i)$ is that it is completely agnostic to the image and only depends on the label distribution. To capture image-specific label relevance, we note that the accuracy of annotating an image with a label may strongly depend on the image. For example, some images may have key aspects of the object occluded. We therefore wish to compute the expected reduction in entropy based on the likelihood that a label is correct $\pli$. When an incorrect label is transmitted, we assume here that no information is passed to the listener (there is an interesting research question about negative information value in this case, which is outside the scope of this paper). 
The \textit{\textbf{expected entropy-reduction}} is therefore
\begin{equation*}
    \label{eq:expected-delta-H}
    E(\Delta H) = \pli \Delta H + (1-\pli)\cdot0
\end{equation*}
this expectation is equivalent to a \textit{\textbf{confidence-weighted entropy reduction}} measure:
\begin{equation}
    \label{eq:conf-delta-H}
    \cwdhmath(i) = \pli \left[H(L) - H[L|l_i=true])\right] \quad,
\end{equation}
where $\pli$ is the probability that $l_i$ is correct and $L$ is a random variable that holds the distribution of all labels. We propose that this is a good measure of label information in the context of a corpus.

\subsection{Other measures of informative labels}
Confidence-weighted ($cw$) entropy reduction, \eqref{eq:conf-delta-H}, is an intuitive quantification of label informativeness, but other properties of the label distribution may capture aspects of label importance. We now discuss two such measures: information about images, and probabilistic surprise. 

\paragraph{Information about images.}
Informative labels were studied in the context of an \textit{image reference game}. In this setup, a speaker provides labels about an image, and a listener needs to identify the target image among a set of distractor images. Recent versions used natural language captioning for the same purpose \cite{andreas2016reasoning,vedantam2017context}.

It is natural to define entropy-reduction for that setup. Similar to \ref{eq:delta-H}, compute the difference between the full entropy over images, and the entropy after transmitting a label. When the distribution over images is uniform the entropy reduction is simply $\log(\textrm{num. images}) - \log(\textrm{num. matching images})$, where the second term is the number of images annotated by a label. Considering the confidence of a label we obtain

\begin{equation}
    \label{eq:image-mi}
    cw\text{-}Image \Delta H(i) = \pli \left[\log(q(l_i)\right] \quad,
\end{equation}

where $\pli$ is again the probability that $l_i$ is correct and $q(l_i)$ is the fraction of images with the label $i$. This measure is fundamentally different from \eqref{eq:conf-delta-H} in that it focuses on the distribution of labels over \textit{images}, not their on joint distribution. 

\paragraph{Probabilistic surprise.}
Transmitting a label changes the label distribution, the ``belief" of the listener. This change can be quantified by the Kullback-Leibler divergence between the label distribution with and without transmission.

\begin{equation}
    \label{eq:conf-dkl}
    cw\text{-}D_{KL}(i)\!=\! D_{KL}\!\left(p(\lll|l_i=\textrm{true}) || \plll \right).
\end{equation}

We can use this measure as a scoring function to rank labels by how strongly they affect the distribution.

\paragraph{Entropy reduction in a singleton model.}
Equation (\ref{eq:H-naive}) computes the entropy over the full joint distribution. An interesting approximation of the joint distribution is provided by the singleton model, which models the joint distribution as the product of the marginals $\plll = \prod_i p(l_i)$.
Given this probabilistic model, the joint entropy of all labels is simply the sum of per-label entropies. The reduction of entropy by a transmitted label is simply the entropy of that label. 

For labels that are rare ($p<0.5$) the entropy grows monotonically with $p$. This means that if all labels are rare, then ranking labels by their frequency in the data, yields the same order as when ranking labels by their singleton entropy reduction.

\subsection{Entropy reduction in large label spaces}
Given a corpus of images, we wish to compute the joint distribution of label co-occurrence in an image $p(l_1,\ldots, l_d)$. The scoring functions described above assume that we can estimate and represent the joint distribution over labels. Unfortunately, even for a modest vocabulary size $d$, the distribution cannot be estimated in practice since it has $2^d$ parameters. Instead, we approximate the label distribution using a probabilistic graphical model called a \textbf{Chow-Liu tree} \cite{1054142}. We first describe the graphical model, and then how it is learned from data.

As any probabilistic graphical model, A Chow-Liu tree has two components: First, a tree $G(V,E)$ with $d$ nodes and $d-1$ edges, where the nodes $V$ correspond to the $d$ labels, and the edges $E$ connect the nodes to form a fully-connected tree. The tree is \textit{directed}, and each node $l_i$, except a single root node, has a single parent node $l_j$. 

As a second component, every edge in the graph, connecting nodes $i$ and $j$ is accompanied by a conditional distribution, $p(l_i | parent(l_i))$. Note that this conditional distribution involves only two binary variables, namely a total of 4 parameters. The full model therefore has only $O(d)$ parameters and can be estimate efficiently from data. With these two components, the Chow-Liu model can be used to represent a joint distribution over all labels, which factorizes over the graph 
\begin{equation}
    \label{eq:chow-liu}
    \log\hat{p}(l_1,\ldots, l_d) = \sum_{i=1}^{d} \log p\left(l_{i}| l_{parent(i)}\right) .
\end{equation}
While any tree structure can be used to represent a factored distribution as in \eqref{eq:chow-liu}, the Chow-Liu algorithm finds one specific tree structure: The distribution that is closest to the original full distribution terms of the Kullback-Liebler divergence $D_{KL}(\hat{p(L)}||p(L) )$. That tree is found in two steps: First, for every pair of labels $i$, $j$, compute their $2\times2$ joint distribution in the image corpus, then compute the mutual information of that distribution. 
\newcommand{\pij}{p_{ij}}
\newcommand{\pii}{p_i}
\newcommand{\pj}{p_j}
\begin{equation}
    MI_{ij} = \sum_{l_i=T,F}\sum_{l_j=T,F} \pij(l_i, l_j) \frac{\pij(l_i, l_j)}{\pii(l_i)\pj(l_j)}
    \label{eq:mi}
\end{equation}
where the summation is over all combination of True and False value for the two variables, $\pij$ is the joint distribution over label co-occurrence, and $\pii$ and $\pj$ are the marginals of that distribution. 

\begin{table*}[th]
\begin{center}
    \begin{tabular}{l|cccccc} \hline
    \toprule
        \multirow{2}{*}{\includegraphics[height=0.43in]{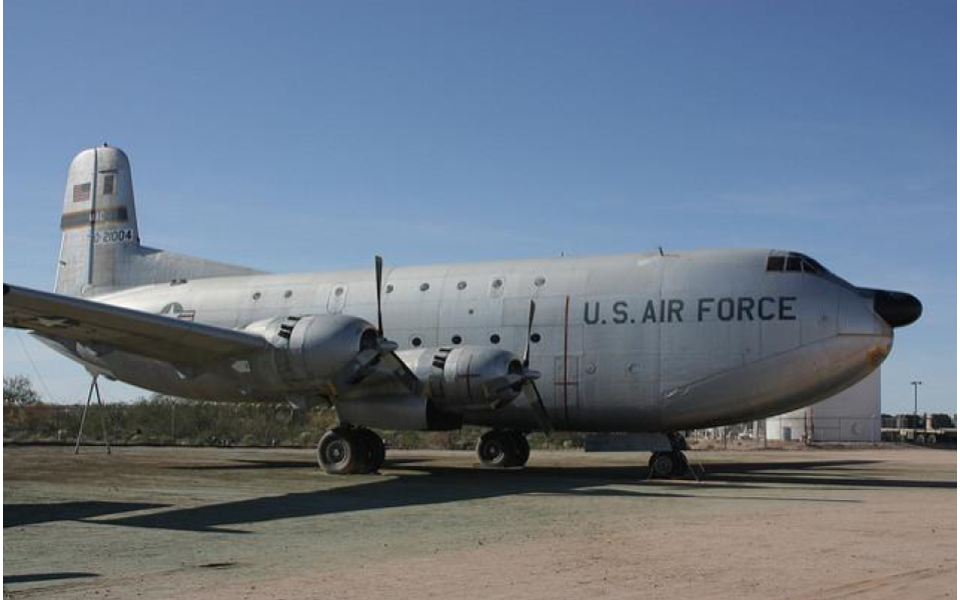}} 
        &\multicolumn{1}{c|}{} & \multicolumn{1}{c|}{}&\multicolumn{1}{c|}{}&  \multicolumn{1}{c|}{} & \multicolumn{1}{c|}{}& \multicolumn{1}{c}{} \\ \cline{2-7} 
        & \multicolumn{1}{c|}{confidence} & \multicolumn{1}{c|}{\cwdh}&\multicolumn{1}{c|}{cw-$D_{KL}$}&  \multicolumn{1}{c|}{\cwmidh{}} & \multicolumn{1}{c|}{cw-$p(x)$}& \multicolumn{1}{c}{cw-Singleton} \\ \cline{2-7}
        &\multicolumn{1}{c|}{vehicle, airplane}& \multicolumn{1}{c|}{airplane} & \multicolumn{1}{c|}{airliner} & \multicolumn{1}{c|}{airliner}&\multicolumn{1}{c|}{vehicle}&\multicolumn{1}{c}{vehicle} \\ 
        \hline
        airplane          &	\textbf{1.0}	& \textbf{52.18}&   56.65	      & 5.71	      & 0.019	&   0.14\\ 
        airline     	  & 0.9	        & 47.53	        & 57.40	    & 5.94	&   0.009	&   0.07\\
        airliner	      & 0.9	        & 46.69	        & \textbf{58.36}&   \textbf{6.29}	&   0.007	&   0.06\\
        aircraft          & 0.9     	& 46.54	        & 46.67	    & 4.83	&   0.022	&   0.15\\
        vehicle	          & \textbf{1.0}& 41.02	        & 14.34	    & 2.33	&   \textbf{0.199}	&   \textbf{0.72}\\
        propeller-aircraft	& 0.8 & 41.01 & 49.97	      & 5.85	&   0.005	&   0.04\\
        aviation            & 0.8 & 40.97 & 40.01	      & 4.30	& 0.019	&   0.13\\
        narrow-body aircraft& 0.8 & 40.73 & 55.06	      & 6.17	&   0.004	&   0.03\\
        air force           & 0.6 & 29.61 & 29.34	      & 3.71	&   0.008	&   0.06\\
        aircraft engine	    & 0.6 & 28.14 & 23.51	      & 3.82	&   0.007	&   0.06\\ 
        \bottomrule
    \end{tabular}
\end{center}
\caption{\textbf{Ranking of the classifier's image-annotations by the different scoring--functions}. We rank the annotations of an image based on the scores assigned to the label. The position (namely, k) of the ground-truth label (in bold) is then used to compute precision and recall. Later, precision and recall are averaged across images} \label{table:ranking}
\end{table*}

As a second step, assign $MI_{ij}$ as the weight of the edge connecting the nodes of labels $i$ and $j$ and find the maximum spanning tree on the weighted graph. Note that the particular directions of the edges of the model are not important. Any set of directions that forms a consistent tree (having at most one parent per node), defines the same distribution over the graph \cite{1054142}. In practice, since committing to a single tree may be sensitive to small perturbations in the data, we model the distribution as a mixture of $k$ trees, which are created by a bootstrap procedure. Details are discussed below. 

Representing the joint distribution of labels using a tree provides great computationally benefits, since many properties of the distribution can be computed very efficiently. Importantly, when the joint distribution factorizes over a tree, the entropy can be computed exactly using the entropy chain rule:
\begin{eqnarray}
    \label{eq:H-tree}
    H[p(l_1,\ldots,l_d)] &=& H[\prod_{i=1}^d p(l_i|parent(l_i))] \\ \nonumber
    &=&  \sum_{i=1}^{d} H[p(l_i|parent(l_i))].
\end{eqnarray}
Here we abused the notation slightly, the root node does not have a parent hence its entropy is not conditioned on a parent but should be $H[p(l_{root})]$.

Furthermore, in a tree-structured probabilistic model, 
one can redirect the edges by selecting any node to be a root, and conditioning all other nodes accordingly \cite{koller2009probabilistic}. This allows us to compute the labeled-conditioned entropy using the following steps. 
First, given a new root label $l_i$, iteratively redirect all edges in the tree to make all nodes its descendents.
Update the conditional density tables on the edges.
Second, assign a marginal distribution of $[0,1]$ to the node $l_i$, reflecting the fact that the label is assigned to be true. Third, propagate the distribution throughout the graph using the conditional probability functions on the edges. Finally, compute the entropy of the new distribution using the chain rule. 

\subsection{Selecting labels for transmission}
Given the above model, we can compute the expected entropy reduction for each label for a given image. We then take an information-retrieval perspective, rank the labels by their scores and emit the highest rank label. 

This process can be repeated for transmitting multiple labels. For example, given that label $l_i$ was transmitted first, we compute how much each of the remaining labels reduces the entropy further. Formally, to decide about a second label to transmit, we compute for every label $l_j \ne l_i$: 
\begin{eqnarray}
    \label{eq:delta-Hj}
    \Delta H_i(j) &=& H[p(\lll|l_i\!=\!true)] \\ \nonumber
                  & & - H[p(\lll|l_i\!=\!true, l_j\!=\!true)]
                  \vspace{-10pt}
\end{eqnarray}
Intuitively, selecting a second label that maximizes this score tends to select labels that are semantically remote from the first emitted labels. If a second label (say, $l_j=$\textit{pet}) is semantically similar to the first label (say, $l_i=$\textit{dog}), the residual entropy of \textit{pet} after observing the label \textit{dog} is low, hence the speaker will prefer other labels. 

\begin{table}[H]
    \small
    \centering
        \begin{tabular}{l|cc|cc} 
            & \multicolumn{2}{c|}{single label} & \multicolumn{2}{c}{multiple labels} \\
            \hline 
            & P@1    & R@5 & P@1 & R@1 \\
            & R@1    &     &     &   \\
            \hline
            \textbf{Scoring functions} &&&&\\
            $cw-\Delta{H}$    &\tabbf{64}&\tabbf{96}&\tabbf{63}&\tabbf{57}  \\
            $cw-D_{KL}$     & 0.43 & 0.96 & 0.42 & 0.38 \\
            \cwmidh{}         & 0.28 & 0.78 & 0.33 & 0.30 \\
            $cw-Singleton$    & 0.33 & 0.88 & 0.34 & 0.31 \\
            $cw-p(x)$       & 0.33 & 0.89 & 0.34 & 0.31 \\
            \hline
            \textbf{Baselines} &&&&\\ 
            confidence      & 0.49 & 0.96 & 0.50 & 0.46 \\
            random          & 0.12 & 0.89 & 0.21 & 0.18 \\
            \hline
            \textbf{Non-weighted} &&&& \\
            $\Delta{H}$       & 0.29 & 0.86 & 0.34 & 0.31 \\
            $D_{KL}$        & 0.22 & 0.87 & 0.29 & 0.26 \\
            $Image\Delta{H}$        & 0.14 & 0.64 & 0.21 & 0.18 \\
            $Singleton$       & 0.26 & 0.88 & 0.29 & 0.26 \\
            $p(x)$          & 0.26 & 0.88 & 0.29 & 0.26 \\
            \hline
        \end{tabular}
        \vspace{5pt}
        \caption{\textbf{Precision and recall of the various approaches}. \cwdh\thinspace reach precision of 64\% for predicting a single label and 63\% in a multi-label setup.}
    \label{table:pr_sl}
\end{table} 

\section{Experiments}
\subsection{Data}
We tested IOTA on the open-images dataset (OID) \cite{OpenImages}. In OID, each image is annotated with a list of labels, together with a confidence score. We approximate the joint label distribution over the validation set (41,620 images annotated with 512,093 labels) and also over the test set (125,436 images annotated with 1,545,835 labels).

\textbf{Ground-truth data (OID-IOTA-10K).} we collected a new dataset of ground-truth ``informative'' labels for 10K images: 2500 from OID-validation and 7500 from OID-test, 3 raters per image. Specifically, raters were instructed to focus on the object or scene that is dominant in the image and to avoid overly generic terms that are not particularly descriptive ("a picture"). Labels were entered as free text, and if possible, matched in real time to a predefined knowledge graph (64\% of samples) so raters can verify label meaning. For 36\% of annotations that were not matched during rating, we mapped them as a post-process to appropriate labels. This process included stemming, resolving ambiguities (\textit{e.g.} deciding if a \textit{bat} meant the animal or the sport equipment) and resolving synonyms (\textit{e.g.} pants and trousers). Overall, in many cases raters used exactly the same term to describe an image. In 68\% of the images \textit{at least} two raters described the image with the same label, and in 27\% all raters agreed. We made the data publicly available at \url{https://chechiklab.biu.ac.il/~brachalior/IOTA/}.

\textbf{Label co-occurrence.} OID lists labels whose confidence is above 0.5. All labels with a count of 100 appearances or more were considered when collecting the label distribution, ignoring their confidence. This yielded a vocabulary of 765 labels. 

\subsection{Evaluation Protocol}
For each of the scoring functions derived above (Sec  \ref{sec:scoring_functions}) we ranked all labels predicted to each image. Given this label ranking, we compared top labels with the ground-truth labels collected from raters, and computed the precision and recall for the top-k ranked labels. Precision and recall are usually used with more than one ground-truth item. In our case however, for each image, there was only one ground-truth label: the majority vote across the three raters. As a result, the precision@1 is identical to recall@1. We excluded images that had no majority vote (3 unique ratings, 27.6\% of images). OID provides confidence values in coarse resolution (1 significant digit), hence multiple labels in an image often share the same confidence values. When ranking by confidence only, we broke ties at random. 

We also tested an evaluation setup where instead of a majority label, every label provided by the three raters was considered as ground truth. Precision and recall was computed in the same way. The code is available at \url{https://github.com/liorbracha/iota}

\subsubsection{Clean and noisy evaluation} \label{sec:clean-noisy}
We evaluated our approach in two setups. In the first, \textit{clean evaluation}, we only considered image labels that were verified to be correct by OID raters. Incorrect labels were excluded from the analysis and not ranked by the scoring functions. We also excluded images whose ground truth label was not in the model's vocabulary. 

In the second setup, \textit{noisy evaluation} we did not force any of these requirements. The analysis included incorrect labels as well as images whose ground truth labels were not in the vocabulary; and thus could not be predicted by our model. As expected, the precision and recall in this setting were significantly lower.  

\subsection{Compared scoring functions and baselines}
We compared the following information-theoretic scoring functions, all weighted by classifier confidence. 
\textbf{(1) Entropy-reduction \cwdh \thinspace} \eqref{eq:conf-delta-H} computed over a mixture of 10  Chow-Liu trees. \textbf{(2) Probabilistic surprise $cwD_{KL}$}, \eqref{eq:conf-dkl}
\textbf{(3) Image-$\Delta H$} Entropy reduction about images,  \eqref{eq:image-mi}.  \textbf{(4) Singleton} entropy reduction-- as in \cwdh\thinspace, but computed over a model that ignores cross-label correlations, and treats the joint label distribution as consisting of independent singletons.

We also evaluated three simpler baselines: \textbf{(5) random} A random ranking of labels within each image.  \textbf{(6) confidence}, ranking based on classifier confidence only, where labels with highest confidence were ranked first. When two labels had the same confidence values, we broke ties randomly. \textbf{(7) term frequency $\pli$}, ranked in a descending order. Note that in our data, the term frequency produces the same ranking as singletons, because all labels have a marginal probability below $0.5$, hence monotonically increase with the entropy. 

\begin{figure}[h]
\centering
    \includegraphics[width=.9\linewidth]{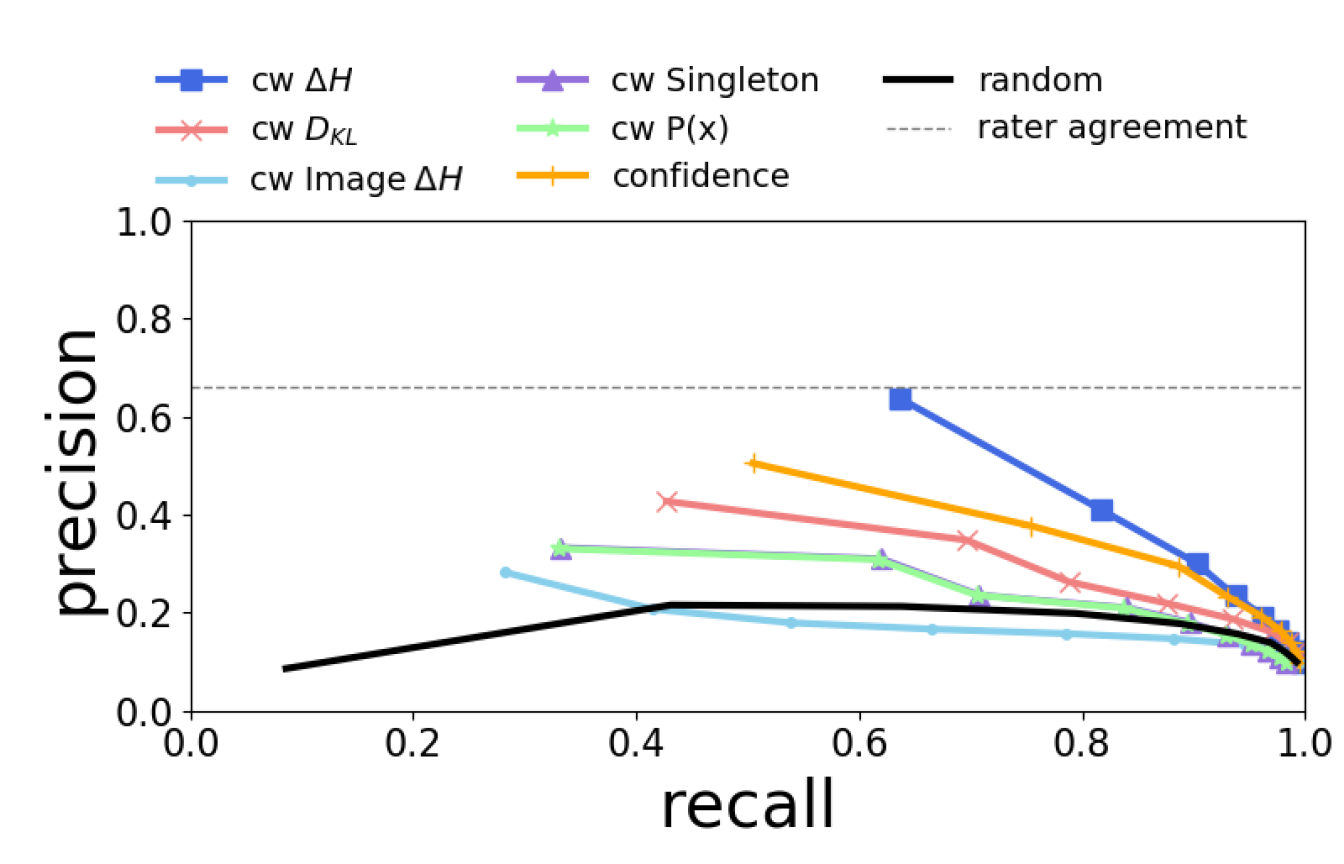}
    \caption{\small \textbf{Precision and recall @k in the \textit{clean} setup}. Results are an averaged over 2877 images from the OID test-set. \cwdh{} (blue curve) achieves precision@1 of 64\% and largely outperforms other scoring functions. Rater agreement (dashed line) is at 66\%, only slightly higher than \cwdh.}
    \label{fig:precision_recall_clean}
    
\end{figure}    
\begin{figure}[h]
\centering
    \includegraphics[width=.9\linewidth]{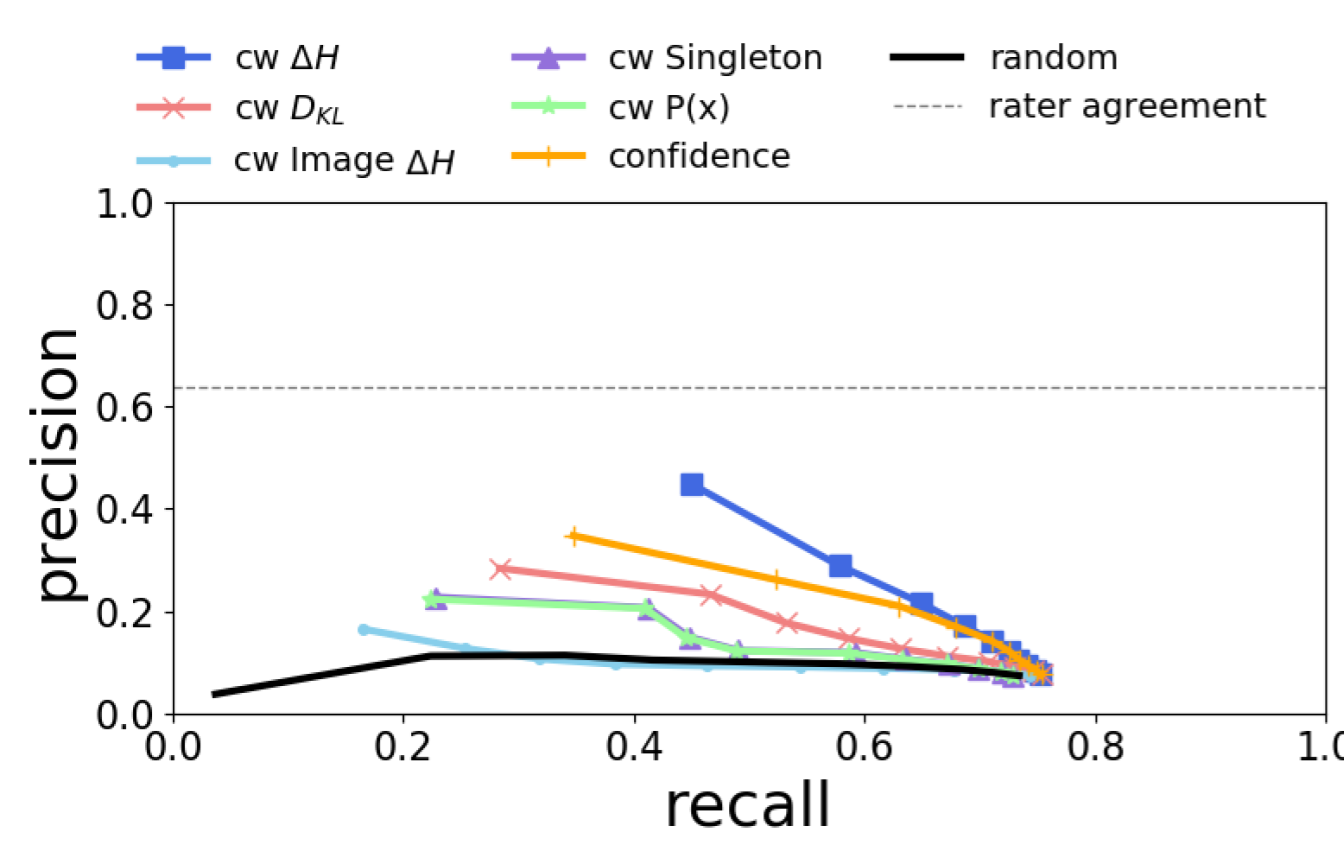}
    \caption{\small\textbf{Precision and recall @k in the \textit{noisy} setup}. Results are an average over 3942 images from the OID test-set. As with the clean set, \cwdh{}(blue curve) outperforms other scoring functions, but only achieves precision@1 of 45\%. Rater agreement (dashed line) is 64\%.}
    \label{fig:precision_recall_noisy}
\end{figure}

\begin{table*}[th]
    \resizebox{\textwidth}{!}{%
    \begin{tabular}{c|c|c|c|c|c|c}
        \toprule
         & Confidence & \cwdh{}  & $cw-D_{KL}$ &\cwmidh{} & $cw-P(x)$ & $cw-Singleton$ \\ \hline\hline
         \includegraphics[height=0.7in]{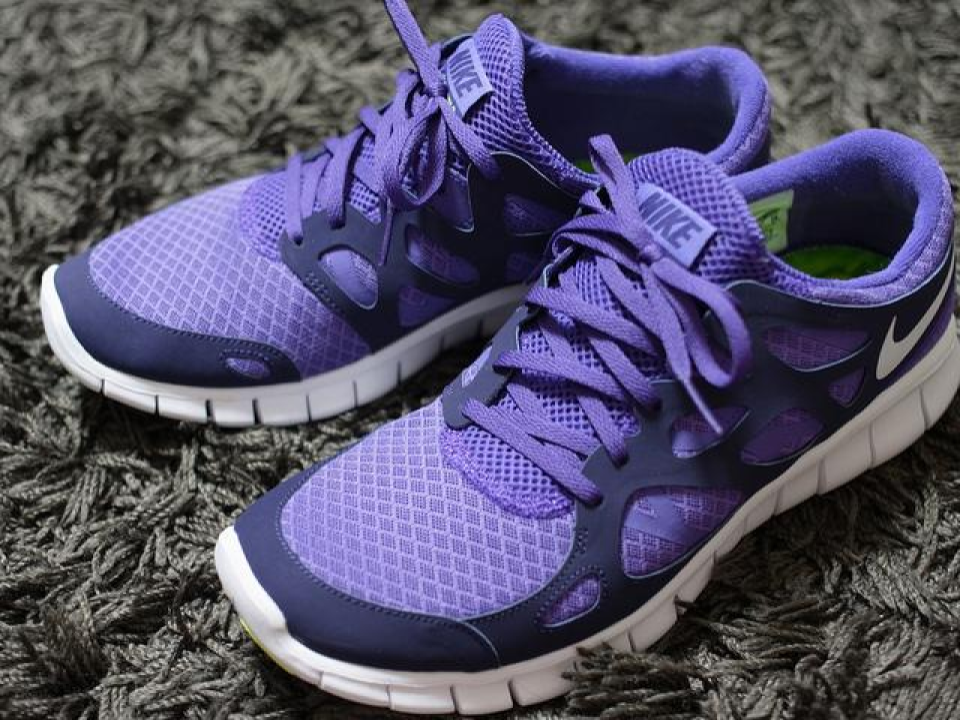} & \textbf{Shoe}, footwear, purple & \textbf{Shoe} & \textbf{Shoe} & Violet & Purple & Purple  \\ \hline
         \includegraphics[height=0.7in]{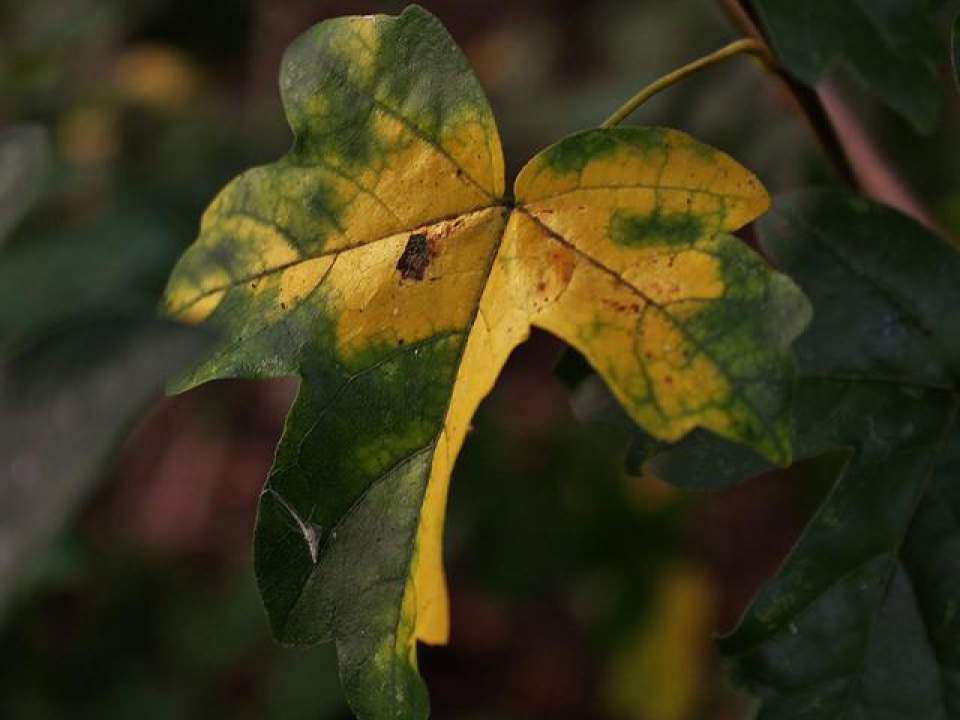} & \textbf{Leaf}, plant, tree, nature, yellow, green & \textbf{Leaf} & Autumn & Season & Land plant & Plant  \\ \hline
         \includegraphics[height=0.7in]{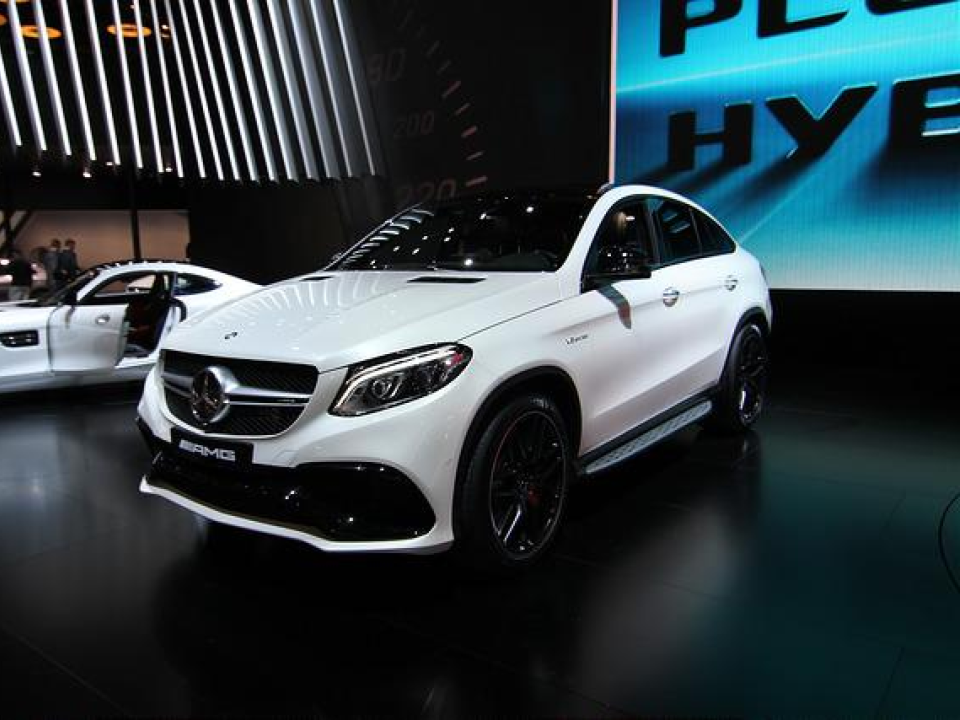} & land vehicle & \textbf{Car}  & Mercedes-benz & Mercedes-benz & Vehicle & Vehicle \\
         \bottomrule
    \end{tabular}
    }
    \vspace{5pt}
    \caption{\textbf{Qualitative example of top-ranked labels by the various scoring functions}. While all annotation are correct \textit{shoe} (top) or \textit{leaf} (middle) are consistent with human annotations. \textit{car} (bottom) $cw-p(x)$ and singletons select an overly abstract label, while $cw-D_{KL}$ and $cw-Image\Delta H$ select more fine grained labels. This effect was pervasive in our dataset.}
    \label{table:examples}
\end{table*}

\section{Results}
We first illustrate label ranking by showing the detailed scores of all scoring functions for one image. Table \ref{table:ranking} annotations (left column) are ordered by \cwdh\thinspace, and the best label per column (scoring function) is highlighted. Note that the classifier gave a confidence of 1.0 to both airplane and vehicle. Singleton and $p(x)$ yield the same ranking (but with different values) because the entropy grows monotonically with $p$. $D_{KL}$ prefers fine-grained classes. 

We next present the precision and recall of IOTA and compared methods over the full OID-test in the \textit{clean} setup (Sec.~\ref{sec:clean-noisy}). Figure \ref{fig:precision_recall_clean} reveals that IOTA achieves high precision, including a p@1 of 64\%. This precision is only slightly lower than the agreement rate of human raters (66\%). See details in Table \ref{table:pr_sl} for comparison. 

Next, we show similar curves for the \textit{noisy} setup (Sec.~\ref{sec:clean-noisy}). Here we also considered images where the ground-truth label is not included in the vocabulary, treating model predictions for these images as false. Figure \ref{fig:precision_recall_noisy} shows that in this case too, \cwdh\thinspace achieves the highest precision and recall compare with the other approaches. As expected, the precision and recall in this setting are lower, reaching precision@1=45\%. 

We further tested all scoring functions using a multilabel evaluation protocol (Sec.~\ref{sec:clean-noisy}). Here, instead of taking the majority label over three rater annotations, we used all three labels (non-weighted) and computed the precision and recall of the scoring functions against that ground truth set. Results are given in Table \ref{table:pr_sl}, showing a similar behavior where \cwdh \thinspace outperforms the other scoring functions. 

\noindent
\textbf{Ablation and comparisons.} 
Several comparisons are worth mentioning. First, confidence- weighted approaches (image-dependent) are consistently superior to non-weighted approaches. This suggests that it is not enough to select "interesting" labels if they are not highly confident for the image. Second, The singleton model (cw-Singleton) performs quite poorly compared to the Chow-Liu tree model (\cwdh). This agrees with our observation that a key factor of label importance is how much it affects uncertainty on other labels. Finally, $Image-\Delta H$, is substantially worse, which is again consistent with the observation that structure in label space is key. 

\begin{figure}[h]
\begin{center}
    \includegraphics[width=1\linewidth]{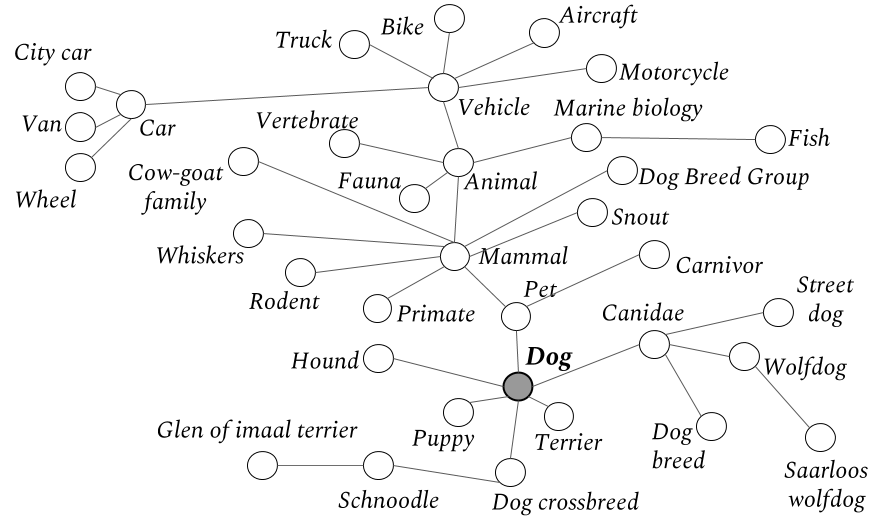}
    \caption{\small \textbf{Part of the Chow-Liu tree around the label \textit{``dog"}.} The model clearly captures semantic relations, even-though they are not explicitly enforced. For instance the label ``\textit{pet}" is connected directly to ``\textit{dog}", and ``\textit{truck}" and ``\textit{bike}" connected to ``\textit{vehicle}"}
\label{fig:tree}    
\end{center}
\end{figure}

\noindent
\textbf{Qualitative Results} 
Table \ref{table:examples} lists top-ranked labels by various scoring functions for three images. \cwdh{} consistently agrees with human annotations (marked in bold), capturing an intermediate, more informative category compared with other scoring functions. In the top row, if based only on high confidence the image content could be described as either \textit{shoe}, \textit{footwear} or \textit{purple}. While all three are technically correct, \textit{shoe} is the most natural, informative title for that image. The middle row (leaf) had 20 predicted annotations (only 6 shown); all approaches other than \cwdh\thinspace failed to return "leaf". Finally, the \textit{car} example (bottom) demonstrates a common phenomena where $cw-P(x)$ and $cw-Singleton$ prefer to more abstract categories whereas $cw-D_{KL}$ and \cwmidh{} prefer fine-grained labels.

These results are all built on a Chow-Liu graphical model. Figure \ref{fig:tree} illustrates parts of the tree that was formed around the label \textit{dog} (38 of 765 labels; validation set). The label-dependency structure reflects sensible label semantics where concepts are grouped in a way that agrees with their meaning (mostly). Note that this tree structure is not a hierarchical model, but only captures the pairwise dependencies among label co-occurrence in the open-images dataset. 

\begin{figure}[h]
      \includegraphics[width=1\linewidth]{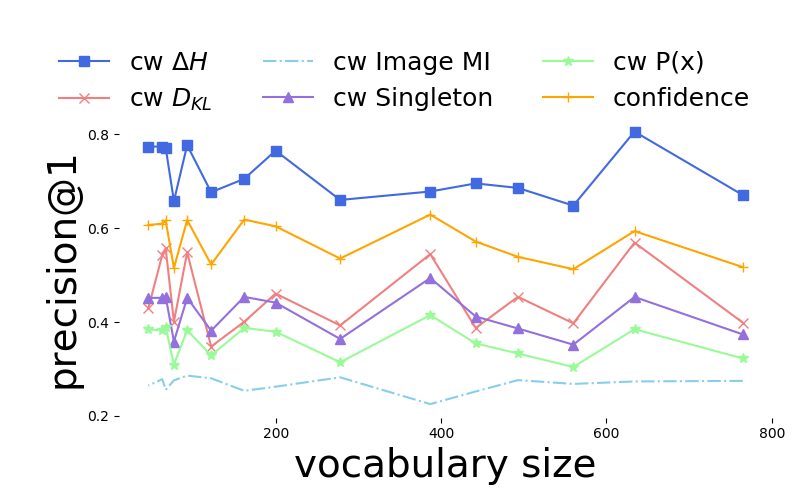}
    \caption{{\small \bf Robustness to vocabulary size}. Different thresholds for the minimum number of label occurrence were tested. The precision of \cwdh{} remains very high for a large range of vocabulary sizes. The relationship between the different scoring functions is consistent as well.}
\label{fig:robust}
\end{figure}

\noindent
\textbf{Robustness to hyper parameters.} 
We tested the robustness of IOTA to the two hyper parameters of the model. (1) The number of trees in the mixture model; and (2) The size of the vocabulary analyzed.

For the first, we computed all scoring functions for tree mixtures with 1,3,5 and 10 trees, and found only a 3\% difference in the p@1 of \cwdh. 

Second, we tested robustness to the number of words in our vocabulary. The vocabulary size is important because our analysis was performed over the most frequent labels in the corpus. As a result, the size of the vocabulary could have affected precision, because entry-level terms (\textit{dog, car}) tend to be more frequent than more fine-grained terms (\textit{e.g.} Labrador, Toyota). We repeated our analysis with different thresholds on the minimum label frequency included in the vocabulary (threshold for values of 50, 100:1000)
Figure \ref{fig:robust} plots the precision@1 of the various scoring functions, showing that the analysis is robust to the size of the vocabulary. 

\section{Conclusions}
We present an unsupervised approach to select informative annotation for a visual scene. We model the prior knowledge of the visual experience using the joint distribution of labels, and use it to rank labels per-image according to how much entropy they can remove over the label distribution. The top ranked labels are the most "intuitive", showing high agreement with human raters. These results are non-trivial as the model does not use any external source of semantic information besides label concurrence.

Several questions remain open. First, while our current experiments capture common context, the approach can be extended to any context. It would be interesting to apply this method to expert annotators with the aim of retrieving listener-specific context. Second, easy-to-learn quantifiers of label importance can be used to improve loss functions in multi-class training, assigning more weight to more important labels. 

\bibliographystyle{unsrt}
\bibliography{ms}

\appendix
\section{Implementation Details}
\label{sec:appendix-algorithm}

Algorithm \ref{alg:model} describes in detail the steps to compute the \cwdh{} scores for a set of labels. Algorithm \ref{alg:eval} describes the inference phase, where the computed scores provide an information-based ranking of the image annotations. Here, we do not specify whether we take a single label as ground truth (by majority) or multiple labels (see Sec 4.2) but give a general framework.

\renewcommand{\algorithmicrequire}{\textbf{Input }}
\renewcommand{\algorithmicensure}{\textbf{Output }}
\begin{algorithm*}[h]
  \caption{\; \cwdh{} scoring-function}
  \label{alg:model}
  \begin{algorithmic}[1]
  \Require ~ \newline (1) A set of image annotations. Each image $i$ annotated with a set of labels $l^i_1,...,l^i_n$, from an open vocabulary. \newline(2) Hyperparameters: number of trees $T$, size $d$ of the output vocabulary.
  \Ensure ~
  \newline (1) Vocabulary of $d$ most common labels ${\cal{L}} = \{l_1,...l_d\}$ 
  \newline (2) \cwdh{} scores for the ${\cal{L}}$
    \item[]
    \For{tree in $1..T$} 
        \State Sub-sample a set of image annotations $A$.
        \State ${\cal{L}} \gets$ $d$ most frequent labels in $A$
        \item[]
        \For{label pair $l_i,l_j \in{\cal{L}}$}
            \State Compute pair (2x2) joint distribution $p(l_i,l_j)$ in $A$
            \State Compute the mutual information $I_{i,j}$ of (Eq. \ref{eq:mi})
        \EndFor
        \item[]
        \State Create Graph $G(V,E)$: $V = {\cal{L}}$, $E$ has weights $I_{ij}$
        \State Find a maximum weight spanning tree (MST, Chow-Liu tree)
        \State Sort the graph such that each node has a single parent
        \State $H \gets$ Compute tree entropy over $G$ (Eq. \ref{eq:H-tree})
        \item[]
        \For{each label $\in{\cal{L}}$}
            \State Set $l_{i}$ as root, direct all other edges such that all node are descendents of $l_{i}$.
            \State Set the root marginal $p(l_{i}) = [0,1]$
            \State Propagate $p(l_{i})$ throughout the tree, compute new marginals. 
            \State $H_i \gets $ label-conditioned entropy of the updated joint distribution
            \State $\Delta{H}(l_{i}) \gets H - H_i$
        \EndFor
    \EndFor
    \State Average $\Delta{H}(l_{i})$ over trees for all labels $(l_{1} ... l_{d})$
  \end{algorithmic}
\end{algorithm*}

\begin{algorithm*}[h]
\begin{algorithmic}[1]
  \caption{\; Ranking and evaluation}
  \label{alg:eval}
    \Require $\Delta{H}(l_{i})$ label scores; ground-truth data
    \Ensure Ranking of image annotations, precision and recall
    \For{image}
        \For{label in ${\cal{L}}$}
        \State    $cw\Delta{H}(l_i) \gets     confidence(l_{i})*\Delta{H}(l_{i})$
        \EndFor
        \State    Rank image annotations by $cw\Delta{H}$. 
        \State    Evaluate against ground-truth label. 
        \State    Compute precision and recall.
    \EndFor
    \State Average precision and recall across images. 
\end{algorithmic}
\end{algorithm*}

\section{Qualitative Examples}
\label{sec:appendix-examples}
\begin{figure*}[ht]
    \centering
    \includegraphics[width=0.99\linewidth]{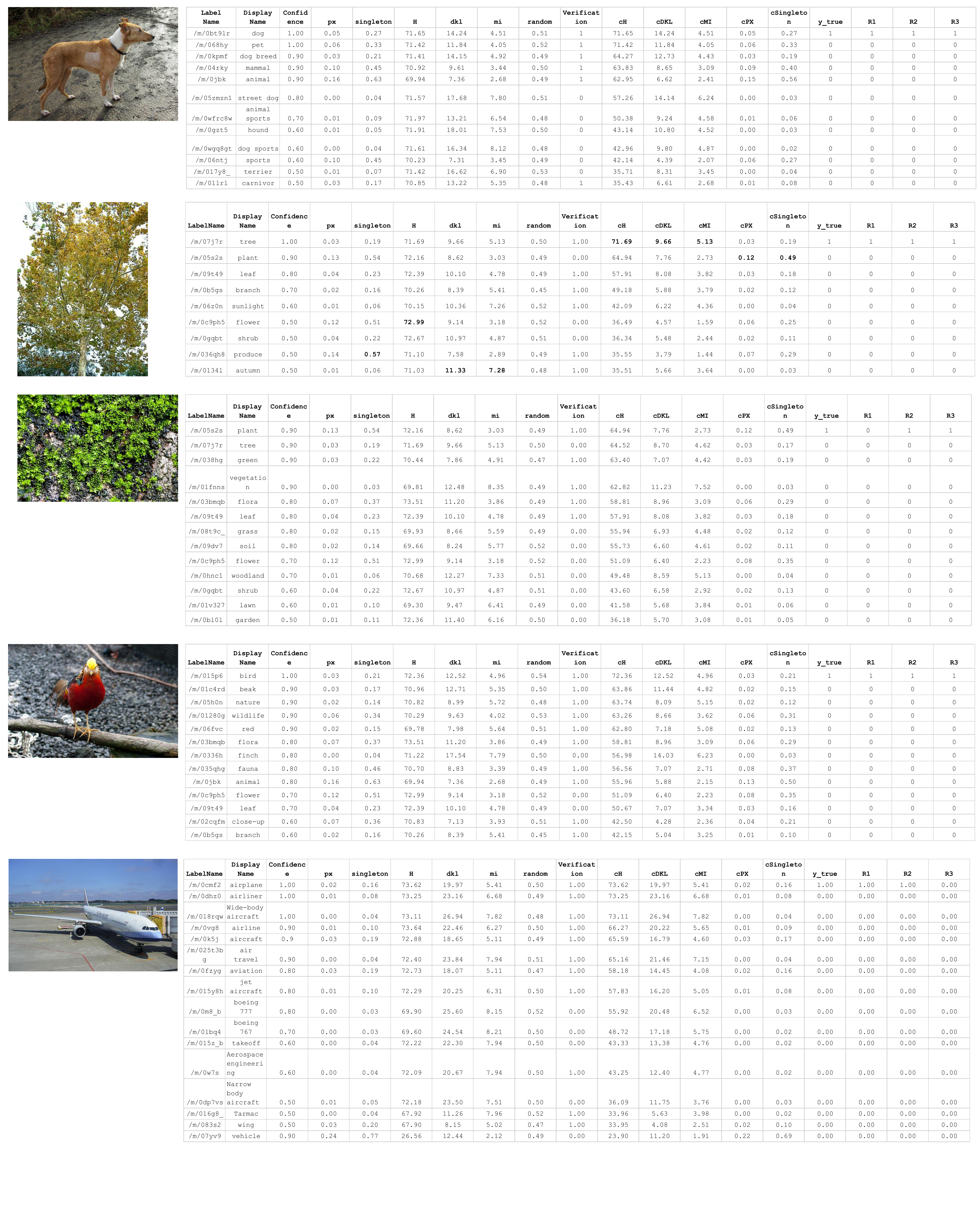}
    \caption{\textbf{Ranked annotations of images from the test set.} The different scoring-functions were analyzed with (right) or without (left) confidence weights. R* mark the raters response (OID-IOTA-10K, see Sec. 4.1).}
    \label{fig:sup_examples}
\end{figure*}

Figure \ref{fig:sup_examples} illustrates the annotations ranking for some images from OID test-set. In these examples we give the full, raw output of our experiments, showing results from all scoring-functions, with or without the confidence weights. "verification" column specifies whether the label was verified by OID raters as correct. "R*" columns present our raters response (see Sec. 4.1) and "y-true" column is the ground truth determined by majority. R* columns in which no entry is marked "1", means that the rater's label was not in the vocabulary. 

\end{document}